\documentclass[10pt,twocolumn,letterpaper]{article}

\usepackage{cvpr}
\usepackage{times}
\usepackage{epsfig}
\usepackage{graphicx}
\usepackage{amsmath}
\usepackage{amssymb}

\usepackage{epsfig}
\usepackage{epstopdf}
\usepackage{makecell,multirow,diagbox}
\usepackage{color}
\usepackage{times}
\usepackage{soul}
\usepackage{url}
\usepackage{amsmath}
\usepackage{array}

\DeclareRobustCommand{\brkbinom}{\genfrac(){0pt}{}}

\cvprfinalcopy %

\usepackage{caption}
\usepackage{xspace}

\def\BeCan{{ABCNet}\xspace}

\usepackage{hyperref}
\hypersetup{
    colorlinks=true,
    linkcolor=blue,
    filecolor=magenta,
    urlcolor=blue,
}

\ifcvprfinal\fi

\begin{document}

\title{\BeCan:
            Real-time Scene Text Spotting with
            Adaptive Bezier-Curve Network\thanks{YL and HC contributed equally to this work.
            YL's contribution was made when visiting
            The University of Adelaide.
            CS is the corresponding author, e-mail: $ \tt chunhua.shen@adelaide.edu.au$}
}
\author{
Yuliang Liu$ ^{\ddag\dag}$
, ~~~ Hao Chen$ ^\dag$,
~~~
Chunhua Shen$ ^\dag$,
~~~
Tong He$ ^\dag$,
~~~
Lianwen Jin$ ^\ddag$,
~~~
Liangwei Wang$ ^\diamond$
\\[0.125cm]
$ ^\ddag$South China University of Technology
~
$ ^\dag$University of Adelaide, Australia
~
$ ^\diamond$Huawei Noah's Ark Lab
}

\maketitle

\begin{abstract}

Scene text detection and recognition has received increasing research attention. Existing methods can be roughly categorized into two groups: character-based and segmentation-based. These methods either are costly for character annotation or need to maintain a complex pipeline, which is often not suitable for real-time applications.
Here
we address the problem by proposing the Adaptive Bezier-Curve Network (\BeCan). Our contributions are three-fold: 1) For the first time, we adaptively fit arbitrarily-shaped text by a parameterized Bezier curve. 2) We design a novel BezierAlign layer for extracting accurate convolution features of a text instance with arbitrary shapes, significantly improving the precision compared with previous methods. 3) Compared with standard bounding box detection, our Bezier curve detection introduces negligible computation overhead, resulting in
superiority of our method in both efficiency and accuracy.

Experiments on arbitrarily-shaped benchmark datasets, namely Total-Text and CTW1500, demonstrate that \BeCan  achieves state-of-the-art accuracy,
meanwhile significantly improving the speed. In particular, on Total-Text, our real-time version is over 10 times faster than recent state-of-the-art methods with a competitive recognition accuracy.

Code is available in the package
\href{https://tinyurl.com/AdelaiDet}{AdelaiDet}.

\end{abstract}

\section{Introduction}
Scene text detection and recognition has received increasing attention due to its numerous applications in computer vision.
Despite
tremendous
progress has been made
recently
\cite{he2016text,wang2019efficient,
liu2019omnidirectional,
shi2016end,liu2019curved,wang2019arbitrary%
},
detecting and recognizing text in the wild remains
largely unsolved
due to its diversity patterns in sizes, aspect ratios, font styles, perspective distortion, and shapes. Although the emergence of deep learning has  significantly improved the performance of the task of scene text spotting, current methods still exist a considerable gap for real-world applications, especially in terms of efficiency.

\begin{figure}[t!]
   \begin{minipage}[c]{0.49\linewidth}
     \centering
     \centerline{\includegraphics[width = 4.cm]{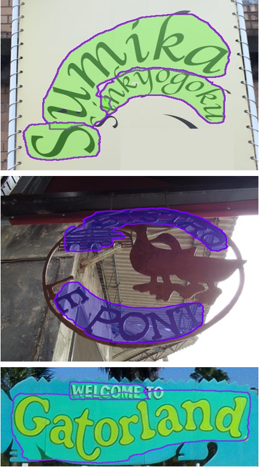}}\label{fig:det_c1}
     \centerline{\small{(a) Segmentation-based method. }}\medskip
   \end{minipage}
   \begin{minipage}[c]{0.49\linewidth}
     \centering
     \centerline{\includegraphics[width = 4.cm]{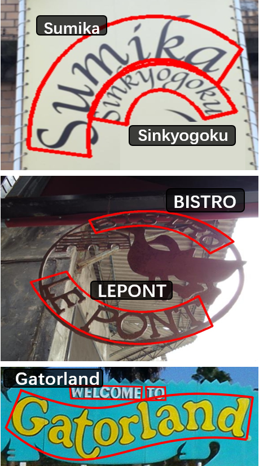}}\label{fig:det_c2}
     \centerline{\small{(b) Our proposed \BeCan. }}\medskip
   \end{minipage}
   \caption{Segmentation-based results are easily affected by nearby text.
   The nonparametric non-structured segmentation results make them very difficult to align features for
   the
   subsequent
   recognition branch.
   Segmentation-based results usually need complex post-processing,
   hampering
   efficiency. Benefiting from the parameterized Bezier curve
   representation, our
   \BeCan  can produce structured detection regions and thus the BezierAlign sampling process can be used for naturally connecting the recognition branch. }\label{fig:intro}
\end{figure}

\begin{figure*}[!t]
   \centering
   \centerline{\includegraphics[width=17.6cm,
   height = 6.7cm]{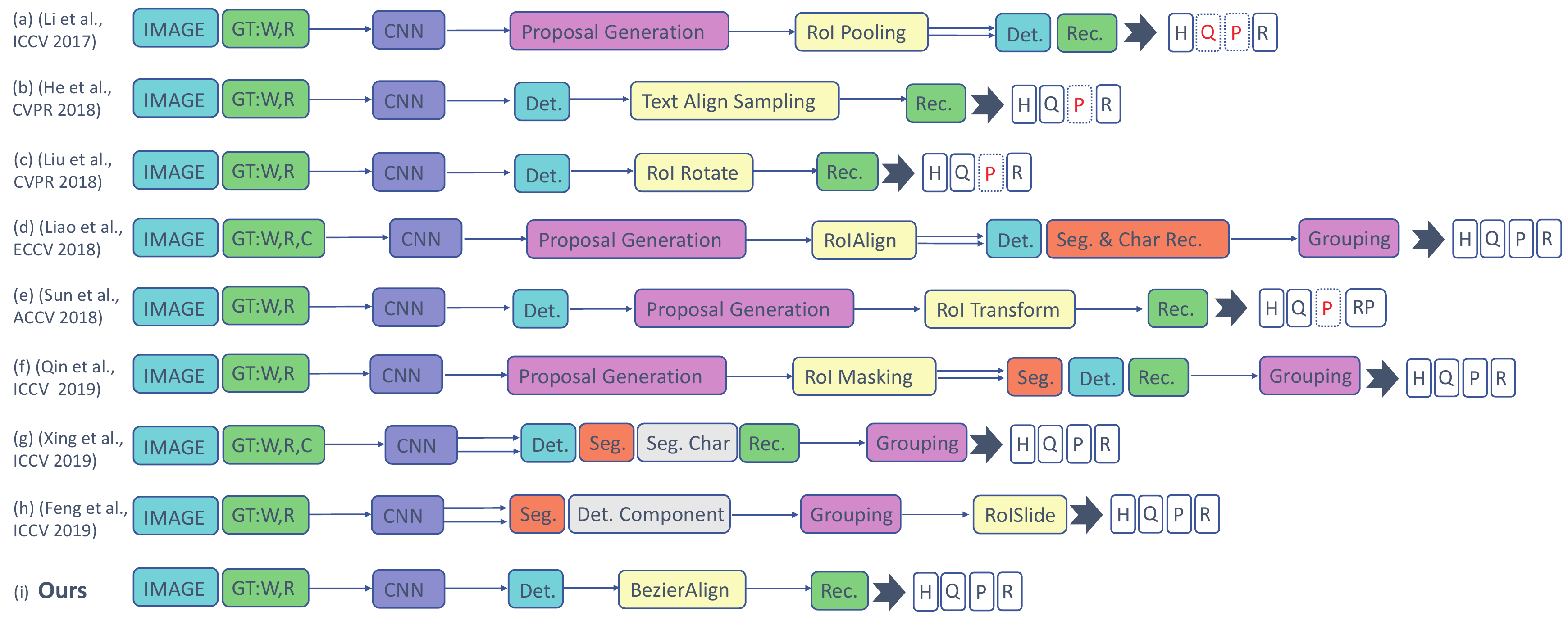}}
   \caption{%
   Overview
   of some end-to-end scene text spotting methods that are most relevant to ours.
   Inside the GT (ground-truth) box, `\textbf{W}', `\textbf{R}', and `\textbf{C}' represent word-level annotation, text content, and character-level annotation, respectively.
   `\textbf{H}', `\textbf{Q}', and `\textbf{P}' represent that the method is able to detect horizontal, quadrilateral, and arbitrarily-shaped text, respectively.
   `\textbf{RP}' means that the method can recognize the curved text inside a quadrilateral box.
   `\textbf{R}': recognition; `\textbf{BBox}': bounding box. Dashed box represents the shape of the text which the method is unable to detect.}\label{fig:related_work}
 \end{figure*}

Recently, many end-to-end methods \cite{lyu2018mask,sun2018textnet,qin2019towards,xing2019convolutional,feng2019textdragon,liao2019mask} have significantly improved the performance of arbitrarily-shaped scene text spotting. However, these methods either use segmentation-based approaches that maintain a complex pipeline or require a large amount of expensive character-level annotations. In addition, almost all of these methods are  slow in inference, hampering the deployment to real-time applications.
Thus,
our motivation is to design a \textit{simple yet effective} end-to-end framework for spotting
oriented or curved
scene text in images \cite{%
ch2019total,liu2019curved}, which ensures fast inference time while achieving an {\it on par} or even better performance compared with state-of-the-art methods.

To achieve this goal, we propose the Adaptive Bezier Curve Network (\BeCan), an end-to-end trainable framework, for arbitrarily-shaped scene text spotting.
\BeCan enables arbitrarily-shaped scene text detection with simple yet effective Bezier curve adaptation, which introduces negligible computation overhead compared with standard rectangle bounding box detection. In addition, we design a novel feature alignment layer---BezierAlign---to precisely calculate convolutional features of text instances in
curved
shapes, and thus high recognition accuracy can be achieved with
almost
negligible computation overhead.
For the first time, we represent the
oriented or curved
text with parameterized Bezier curves, and the results show the effectiveness of our method. Examples of the our spotting results are shown in Figure \ref{fig:intro}.

Note that previous methods such as TextAlign \cite{he2018end} and FOTS \cite{liu2018fots} can be viewed as a special case of \BeCan because a quadrilateral bounding box can be seen as the simplest arbitrarily-shaped bounding box with 4 straight boundaries. In addition, \BeCan can avoid complicated transformation such as 2D attention \cite{li2019towards}, making the design of the recognition branch
considerably simpler.

We summarize our main contributions as follows.
\begin{itemize}
\itemsep -0.1cm
   \item In order to accurately localize
   oriented and curved
   scene text in images, for the first time, we introduce a new  concise parametric representation of
   curved
   scene text using Bezier curves. It introduces negligible computation overhead compared with
   the standard
   bounding box representation.

   \item We propose a sampling method, a.k.a.\ BezierAlign, for accurate feature alignment, and thus the recognition branch can be naturally connected to the overall structure.
   By sharing backbone features, the recognition branch can be designed with a %
   light-weight structure.

   \item The simplicity of our method allows it to perform inference in real time. \BeCan achieves state-of-the-art performance on two challenging datasets, Total-Text and CTW1500,
   demonstrating
   advantages in both effectiveness and efficiency.
\end{itemize}

\begin{figure*}[!t]
   \centering
   \centerline{\includegraphics[width=16.6cm,
   ]{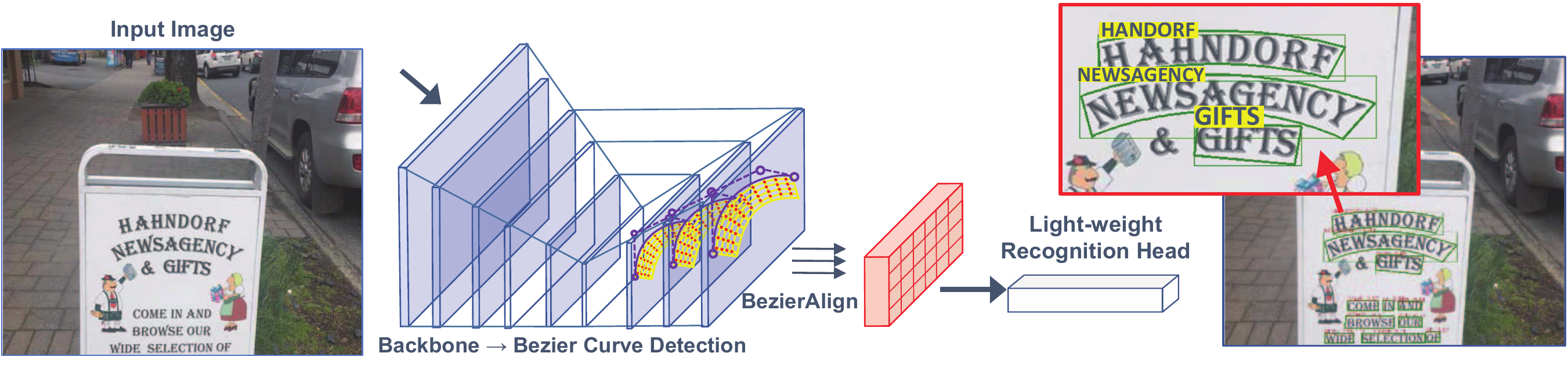}}
   \caption{The framework of the proposed \BeCan. We use cubic Bezier curves and BezierAlign to extract %
   curved
   sequence features  using  the Bezier curve detection results. The
   overall
   framework is end-to-end trainable with  high efficiency. Purple dots represent the control points of the cubic Bezier curve. }\label{fig:pipeline}
 \end{figure*}

\subsection{Related Work}
Scene text spotting requires detecting and recognizing text simultaneously instead of concerning only one task.
Recently, the emergence of deep-learning-based methods have significantly  advanced the performance of text spotting.
Both the detection and recognition have been dramatically improved in performance. We summarized several representative  deep-learning-based scene text spotting methods into the following two categories. Figure \ref{fig:related_work} shows an overview of typical works.

\textbf{Regular End-to-end Scene Text Spotting}
Li et al.\ \cite{li2017towards} propose the first deep-learning  based end-to-end trainable scene text spotting method. The method successfully uses a RoI Pooling \cite{ren2015faster} to joint detection and recognition features via a two-stage framework, but it can only spot  horizontal and focused text. Its improved version \cite{li2019towards} significantly improves the performance, but the speed is limited.
He et al.\ \cite{he2018end} and Liu et al.\ \cite{liu2018fots} adopt an anchor-free mechanism to improve both the training and inference speed. They use a similar sampling strategy, i.e., Text-Align-Sampling and RoI-Rotate, respectively, to enable extracting feature from quadrilateral detection results.
Note that both of these two methods are not capable of spotting arbitrarily-shaped scene text.

\textbf{Arbitrarily-shaped End-to-end Scene Text Spotting}
To detect arbitrarily-shaped scene text, Liao et al.\  \cite{lyu2018mask} propose a Mask TextSpotter which subtly refines Mask R-CNN %
and uses character-level supervision to simultaneously detect and recognize characters and instance masks. The method significantly improves the performance of spotting arbitrarily-shaped scene text.
However, the character-level ground truths are expensive, and using free synthesized data is hard to produce character-level ground truth for real data in practice. Its improved version \cite{liao2019mask} significantly alleviated the reliance for the character-level ground truth. The method relies on a region proposal network, which restricts the speed to some extent. Sun et al.\  \cite{sun2018textnet} propose the TextNet which produces quadrilateral detection bounding boxes in advance, and then use a region proposal network to feed the detection features for recognition. Although the method can directly recognize the arbitrarily-shaped text from a quadrilateral detection, the performance is still limited.

Recently, Qin et al.\  \cite{qin2019towards} propose to use a RoI Masking to focus on the arbitrarily-shaped text region. However, the results may easily be affected by outlier pixels. In addition, the segmentation branch increases the computation burden; the fitting polygon process also introduces extra time consumption; and the grouping result is usually jagged and not smooth. The work in \cite{xing2019convolutional} is the first one-stage arbitrarily-shaped scene text spotting method, requiring character-level ground truth data for training. Authors of \cite{feng2019textdragon} propose a novel sampling method, RoISlide, which uses fused features from the predicting segments of the text instances, and thus it is robust to long arbitrarily-shaped text.

\section{Adaptive Bezier Curve Network (\BeCan)}
\BeCan is an end-to-end trainable framework for spotting arbitrarily-shaped scene text. An intuitive pipeline can be seen in Figure \ref{fig:pipeline}. Inspired by \cite{zhong2019anchor,tian2019fcos,he2017deep}, we adopt a single-shot, anchor-free convolutional neural network as the detection framework. Removal of anchor boxes significantly simplifies the detection for our task. Here the detection is densely predicted on the output feature maps of the detection head, which is constructed by 4 stacked convolution layers with stride of 1, padding of 1, and 3$\times$3 kernels.
Next,
we present the key components of the proposed \BeCan in two parts: 1) Bezier curve detection; and 2) BezierAlign and recognition branch.

\subsection{Bezier Curve Detection}
\label{subsec:Bezier_det}
Compared to segmentation-based methods \cite{wang2019shape,xu2019textfield,baek2019character,tian2019learning,
zhang2019look,long2018textsnake}, regression-based methods are more direct solutions
to arbitrarily-shaped text detection, e.g., \cite{liu2019curved,wang2019arbitrary}. However, previous regression-based methods require complicated parameterized prediction to fit the text boundary, which is not very efficient and robust for the various text shapes in practice.

\begin{figure*}[!t]
   \centering
   \centerline{\includegraphics[width=0.7\textwidth,
   ]{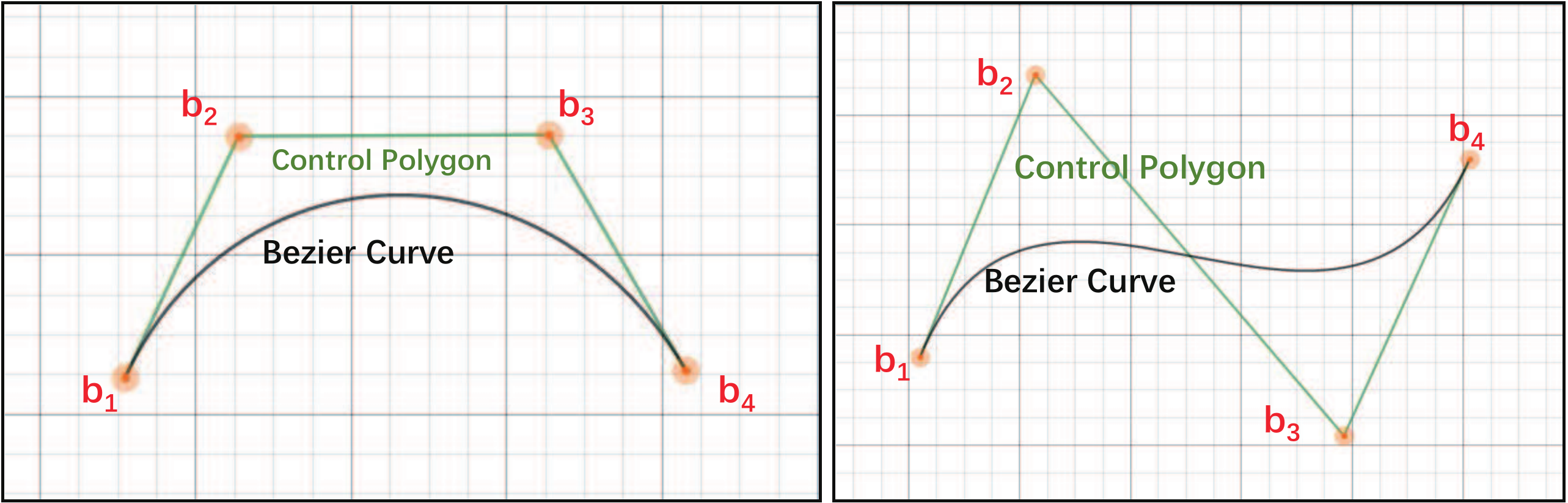}}
   \caption{
   Cubic Bezier curves. $b_{i}$ represents the control points. The green lines forms a control polygon, and the black curve is the cubic Bezier curve.
   Note that with only two end-points $ b_1 $ and $b_4 $ the Bezier curve degenerates to a straight line.
   }
\label{fig:Bezier_ill}
\end{figure*}

To simplify the arbitrarily-shaped scene text detection, following the regression method, we believe that the Bezier curve is an ideal concept for  parameterization of curved text.
The Bezier curve represents a parametric curve  $c(t)$ that uses the Bernstein Polynomials \cite{lorentz2013bernstein} as its basis. The definition is shown in Equation \eqref{eq:Bezier_def}.
\begin{equation}\label{eq:Bezier_def}
   c(t) = \sum_{i=0}^{n}b_{i}B_{i,n}(t), 0\leq t\leq 1,
 \end{equation}
where, $n$ represents the degree, $b_{i}$ represents the $i$-$th$ control points, and $B_{i,n}(t)$ represents the Bernstein basis polynomials, as shown in Equation \eqref{eq:Bernstein}:
\begin{equation}\label{eq:Bernstein}
   B_{i,n}(t) = \brkbinom{n}{i}t^{i}(1-t)^{n-i}, i=0,...,n,
 \end{equation}
where $\brkbinom{n}{i}$ is a binomial coefficient. To fit arbitrary shapes of the text with Bezier curves, we comprehensively observe arbitrarily-shaped scene text from the existing datasets and the real world, and we empirically show that a cubic Bezier curve (i.e., $n$ is 3) is sufficient to fit different kinds of the arbitrarily-shaped scene text in practice. An illustration of cubic Bezier curve is shown in Figure \ref{fig:Bezier_ill}.

Based on the cubic Bezier curve, we can simplify the arbitrarily-shaped scene text detection to a bounding box regression with eight control points in total. Note that a straight text that has four control points (four vertexes) is a typical case of arbitrarily-shaped scene text. For consistency, we interpolate additional two control points in the tripartite points of each long side.

To learn the coordinates of the control points, we first generate the Bezier curve ground truths described in \ref{subsubsec:Bezier_gen} and follow a similar regression method as in \cite{liu2017deep} to regress the targets. For each text instance, we use
\begin{equation}\label{eq:regress}
   \Delta_{x} = b_{ix}-x_{min}, \Delta_{y} = b_{iy}-y_{min},
 \end{equation}
where $x_{min}$ and $y_{min}$ represent the minimum $x$ and $y$ values of the 4 vertexes, respectively. The advantage of predicting the relative distance is that it is irrelevant to whether the Bezier curve control points are beyond the image boundary. Inside the detection head, we only need one convolution layer with 16 outputted channels to learn the $\Delta_{x}$ and $\Delta_{y}$, which is nearly cost-free while the results can still be accurate, which will be discussed in Section \ref{sec:exp}.

\subsubsection{Bezier Ground Truth Generation}
\label{subsubsec:Bezier_gen}
\begin{figure}[t!]
  \begin{minipage}[c]{0.49\linewidth}
    \centering
    \centerline{\includegraphics[width = 4.1cm, height = 3.5cm]{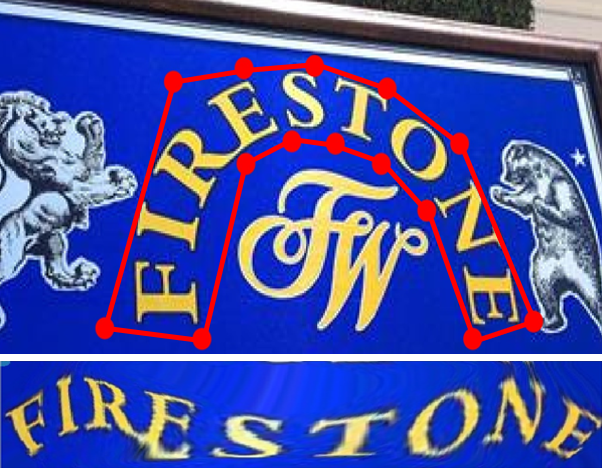}}\label{fig:det_c1}
    \centerline{\small{(a) Original ground truth. }}\medskip
  \end{minipage}
  \begin{minipage}[c]{0.49\linewidth}
    \centering
    \centerline{\includegraphics[width = 4.1cm, height = 3.5cm]{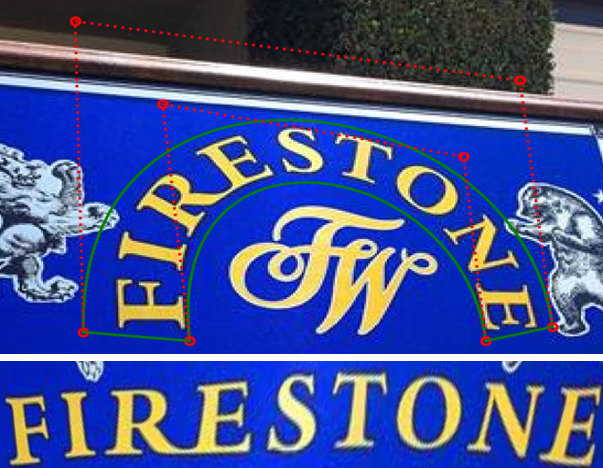}}\label{fig:det_c3}
    \centerline{\small{(b) Generated results. }}\medskip
  \end{minipage}
  \caption{Comparison of Bezier curve generation. In Figure (b), for each curve boundary, the red dash lines form a control polygon, and the red dots represent the control points. Warping results are showed below. In Figure (a), we utilize TPS \cite{bookstein1989principal} and STN \cite{jaderberg2015spatial} to warp the original ground truth into rectangular shape. In Figure (b), we use generated Bezier curves  and our BezierAlign to warp the results.
  }\label{fig:Bezier_gen_comp}
\end{figure}

\begin{figure*}[!t]
   \centering
   \centerline{\includegraphics[width=17.6cm, height = 4cm]{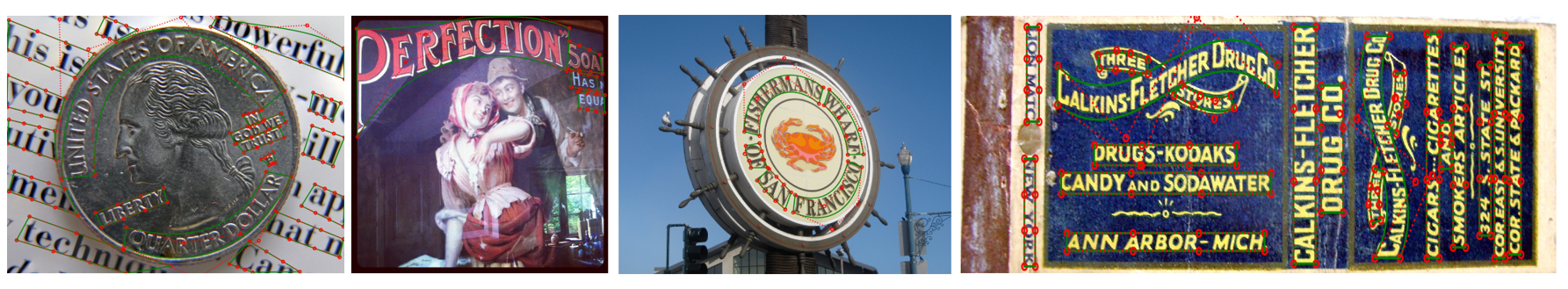}}
   \caption{Example results of Bezier curve generation. Green lines are the final Bezier curve results. Red dash lines represent the control polygon, and the 4 red end points represent the control points. Zoom in for better visualization.}\label{fig:Bezier_gen_examples}
\end{figure*}

In this section, we briefly introduce how to generate Bezier curve ground truth based on the original annotations. The arbitrarily-shaped datasets, e.g., Total-text \cite{ch2019total} and CTW1500 \cite{liu2019curved}, use polygonal annotations for the text regions. Given the annotated points $\{p_{i}\} _{i=1}^{n}$ from the curved boundary, where $p_{i}$ represents the $i$-$th$ annotating point, the main goal is to obtain the optimal parameters for cubic Bezier curves $c(t)$ in Equation \eqref{eq:Bezier_def}. To achieve this, we can simply apply standard least square method, as shown in Equation \eqref{eq:matrix}:
\begin{equation}
\begin{bmatrix}
  B_{0,3}(t_0) & \cdots\ & B_{3,3}(t_0)\\
  B_{0,3}(t_1) & \cdots\ & B_{3,3}(t_1)\\
  \vdots&  \ddots\ &  \vdots\\
  B_{0,3}(t_m) & \cdots\ & B_{3,3}(t_m)
\end{bmatrix}
\begin{bmatrix}
 b_{x_0} & b_{y_0}\\
 b_{x_1} & b_{y_1}\\
 b_{x_2} & b_{y_2}\\
 b_{x_3} & b_{y_3}\\
\end{bmatrix}
=
\begin{bmatrix}
 p_{x_0} & p_{y_0}\\
 p_{x_1} & p_{y_1}\\
 \vdots & \vdots\\
 p_{x_m} & p_{y_m}\\
\end{bmatrix}
\label{eq:matrix}
\end{equation}
Here $m$ represents the number of annotated points for a curved boundary. For Total-Text and CTW1500, $m$ is 5 and 7, respectively. $t$ is calculated by using the ratio of the cumulative length to the perimeter of the polyline. According to Equation \eqref{eq:Bezier_def} and Equation \eqref{eq:matrix}, we convert the original polyline annotation to a parameterized Bezier curve. Note that we directly use the first and the last annotating points as the first ($b_0$) and the last ($b_4$) control points, respectively. An visualization comparison is shown in the Figure \ref{fig:Bezier_gen_comp}, which shows that the generating results can be even visually better than the original ground truth. In addition, based on the structured Bezier curve bounding box, we can easily using our BezierAlign described in Section \ref{subsec:bezieralign} to warp the curved text into a horizontal format without dramatic deformation.
More examples of the Bezier curve generation results are shown in Figure \ref{fig:Bezier_gen_examples}. The simplicity of our method allows it generalize to different kinds of text in practice.

\begin{figure*}[t!]
   \begin{minipage}[c]{0.33\linewidth}
     \centering
     \centerline{\includegraphics[width = 5.5cm, height = 3.5cm]{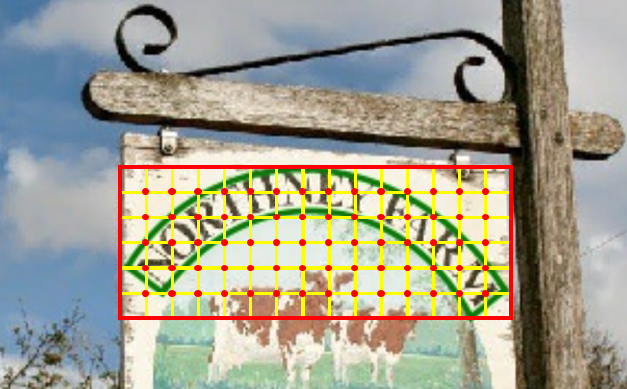}}\label{fig:det_c1}
     \centerline{\small{(a) Horizontal sampling. }}\medskip
   \end{minipage}
   \begin{minipage}[c]{0.33\linewidth}
     \centering
     \centerline{\includegraphics[width = 5.5cm, height = 3.5cm]{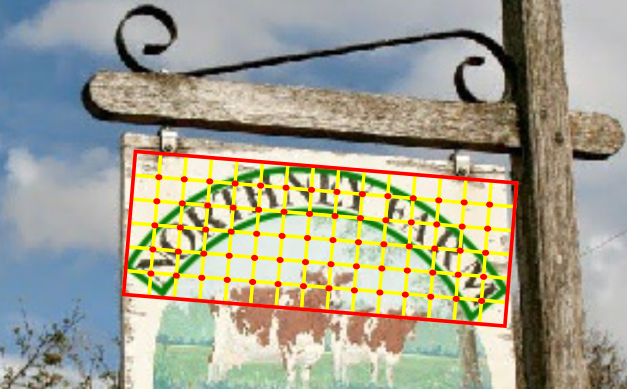}}\label{fig:det_c2}
     \centerline{\small{(b) Quadrilateral sampling. }}\medskip
   \end{minipage}
   \begin{minipage}[c]{0.33\linewidth}
     \centering
     \centerline{\includegraphics[width = 5.5cm, height = 3.5cm]{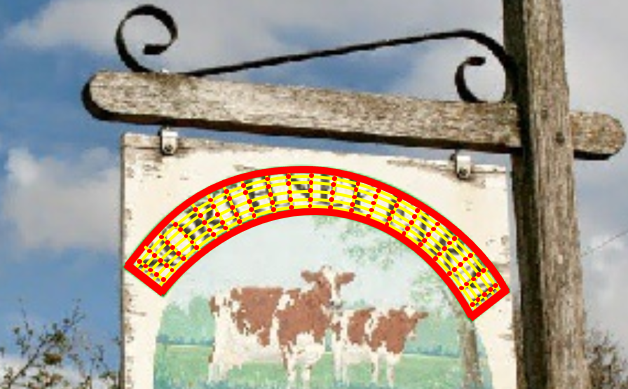}}\label{fig:det_c3}
     \centerline{\small{(c) BezierAlign. }}\medskip
   \end{minipage}

   \caption{Comparison between previous sampling methods and BezierAlign. The proposed BezierAlign can accurately sample features of the text region, which is essential for recognition training. Note that the align procedure is processed in intermediate convolutional features.}\label{fig:Bezier_align}
\end{figure*}

\subsubsection{Bezier Curve Synthetic Dataset}
\label{subsubsec:Bezier_syn}
\begin{figure*}[!h]
   \centering
   \centerline{\includegraphics[
   width=0.7\textwidth
   ]{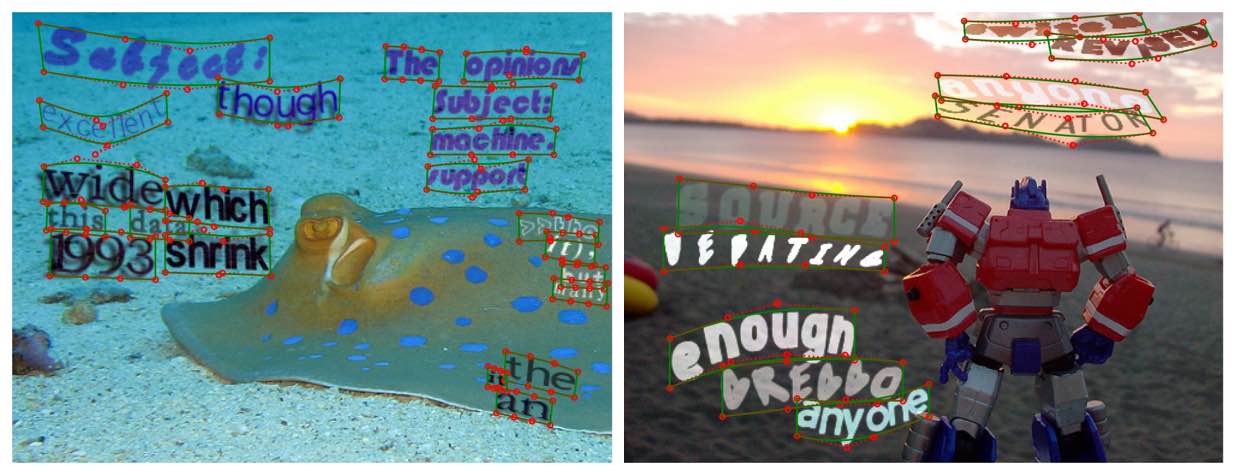}}
   \caption{Examples of cubic Bezier curve synthesized data.}\label{fig:Bezier_syn}
\end{figure*}

For the end-to-end scene text spotting methods, a massive amount of free synthesized data is always necessary, as shown in Table \ref{tab:tt}. However,
the existing 800k SynText dataset \cite{gupta2016synthetic} only provides quadrilateral bounding box for a majority of straight text.
To diversify and enrich the arbitrarily-shaped scene text, we make some effort to synthesize 150k synthesized dataset (94,723 images contain a majority of straight text, and 54,327 images contain mostly curved text) with the VGG synthetic method \cite{gupta2016synthetic}. Specially, we filter out 40k text-free background images from COCO-Text \cite{veit2016coco} and then prepare the segmentation mask and scene depth of each background image with \cite{pont2016multiscale} and \cite{laina2016deeper} for the following text rendering. To enlarge the shape diversity of synthetic texts, we modify the VGG synthetic method by synthesizing scene text with various art fonts and corpus and generate the polygonal annotation for all the text instances.
The annotations are then used for producing Bezier curve ground truth by the generating method described in Section \ref{subsubsec:Bezier_gen}. Examples of our synthesized data are shown in Figure \ref{fig:Bezier_syn}.

\subsection{BezierAlign}
\label{subsec:bezieralign}
To enable end-to-end training, most of the previous methods adopt various sampling (feature alignment) methods to connect the recognition branch.
Typically a sampling method represents an in-network region cropping procedure.
In other words, given a feature map and Region-of-Interest (RoI), using the sampling method to select the features of RoI and efficiently output a feature map of a fixed size.
However, sampling methods of previous non-segmentation based methods, e.g., RoI Pooling \cite{li2017towards}, RoI-Rotate \cite{liu2018fots}, Text-Align-Sampling \cite{he2018end}, or RoI Transform \cite{sun2018textnet} cannot properly align features of arbitrarily-shaped text (RoISlide \cite{feng2019textdragon} numerous predicting segments).
By exploiting the parameterization nature of a compact Bezier curve bounding box, we propose BezierAlign for feature sampling. BezierAlign is extended from  RoIAlign \cite{He2017Mask}.
Unlike RoIAlign, the shape of sampling grid of BezierAlign is not rectangular. Instead, each column of the arbitrarily-shaped grid is orthogonal to the Bezier curve boundary of the text. The sampling points have equidistant interval in width and height, respectively, which are bilinear interpolated %
with respect
to the coordinates.

Formally
given an input feature map and Bezier curve control points, we concurrently process all the output pixels of the rectangular output feature map with size $h_{out}\times w_{out}$. Taking pixel $g_i$ with position $(g_{iw}, g_{ih})$ (from output feature map) as an example, we calculate $t$ by Equation \eqref{eq:Bezier_timestep}:
\begin{equation}\label{eq:Bezier_timestep}
   t = \frac{g_{iw}}{w_{out}}.
 \end{equation}
We then use $t$ and Equation \eqref{eq:Bezier_def} to calculate the point of upper Bezier curve boundary $tp$ and lower Bezier curve boundary $bp$. Using $tp$ and $bp$, we can linearly index the sampling point $op$ by Equation \eqref{eq:Bezier_sampling}:
\begin{equation}\label{eq:Bezier_sampling}
   op = bp  \cdot  \frac{g_{ih}}{h_{out}} + tp \cdot  (1 - \frac{g_{ih}}{h_{out}}).
 \end{equation}
With the position of $op$, we can easily apply bilinear interpolation to calculate the result. Comparisons among previous sampling methods and BezierAlign are shown in Figure \ref{fig:Bezier_align}.

\paragraph{Recognition branch.} Benefiting from the shared backbone feature and BezierAlign, we design a light-weight recognition branch as shown in Table \ref{tab:rec_struct},
for faster execution.
It consists of 6 convolutional layers, 1 bidirectional LSTM \cite{hochreiter1997long} layer, and 1 fully connected layer. Based on the output classification scores, we use a classic CTC Loss \cite{graves2006connectionist} for text string (GT) alignment. Note that during training, we directly use the generated Bezier curve GT to extract the RoI features.
Therefore the detection branch does not affect the recognition branch.
In the inference phase, the RoI region is replaced by the detecting Bezier curve described in Section \ref{subsec:Bezier_det}. Ablation studies in Experimental Section \ref{sec:exp} demonstrate that the proposed BezierAlign can significantly improve the recognition performance.

\begin{table}[!t]
   \centering
   \newcommand{\tabincell}[2]{\begin{tabular}{@{}#1@{}}#2\end{tabular}}
   \small
   \begin{tabular}{c|c|c}
     \hline
     \tabincell{c}{Layers \\ (CNN - RNN)}  & \tabincell{c}{Parameters \\ (kernel size, stride)}  & \tabincell{c}{Output  Size \\ ($n$, $c$, $h$, $w$)} \\
     \hline
     conv layers $\times$4 & (3, 1) & ($n$, 256, $h$, $w$) \\
     conv layers $\times$2 & (3, (2,1)) & ($n$, 256, $h$, $w$) \\
     average pool for $h$ &  -  & ($n$, 256, 1, $w$) \\
     \hline
     Channels-Permute & - & ($w$, $n$, 256) \\
     BLSTM  & - & ($w$, $n$, 512) \\
     FC  & - & ($w$, $n$, $n_{class}$) \\
     \hline
   \end{tabular}
   \caption{Structure of the recognition branch, which is a simplified version of CRNN \cite{shi2016end}. For all convolutional layers, the padding size is restricted to 1. $n$ represents batch size. $c$ represents the channel size. $h$ and $w$ represent the height and width of the outputted feature map, and $n_{class}$ represents the number of the predicted class, which is set to 97 in this paper, including upper and lower cases of English characters, digits, symbols, one category representing all other symbols, and an ``EOF'' of the last category. }\label{tab:rec_struct}
 \end{table}

\section{Experiments}
\label{sec:exp}
We evaluate our method on two recently introduced arbitrarily-shaped scene text benchmarks, Total-Text \cite{kheng2017total} and CTW1500 \cite{liu2019curved}, which also contain a
large amount
of straight text. We also conduct ablation studies on Total-Text to verify the effectiveness of our proposed method.

\subsection{Implemented details}
The backbone of this paper follows a common setting as most of the previous papers, i.e., ResNet-50 \cite{he2016deep} together with a Feature Pyramid Network (FPN) \cite{lin2017feature}. For detection branch, we utilize RoIAlign on 5 feature maps with 1/8, 1/16, 1/32, 1/64, and 1/128 resolution of the input image while for recognition branch, BezierAlign is conducted on three feature maps with 1/4, 1/8, and 1/16 sizes.
The pretrained data is collected from publicly  available English word-level-based datasets, including 150k synthesized data described in Section \ref{subsubsec:Bezier_syn}, 15k images filtered from the original COCO-Text \cite{veit2016coco}, and 7k ICDAR-MLT data \cite{nayef2019icdar2019}. The pretrained model is then finetuned on the training set of the target datasets.
In addition, we also adopt data augmentation strategies, e.g., random scale training, with the short size randomly being chosen from 560 to 800 and the long size being less than 1333; and random crop, which we make sure that the crop size is larger than half of the original size and without any text being cut (for some special cases that hard to meet the condition, we do not apply random crop).

We train our model using 4 Tesla V100 GPUs with the image batch size of 32. The maximum iteration is 150K; and the initialized learning rate is 0.01, which reduces to 0.001 at the 70K$^{\rm th}$ iteration and 0.0001 at 120K$^{\rm th}$ iteration. The whole training process takes about 3 days.

\begin{table*}[!t]
   \centering
   \newcommand{\tabincell}[2]{\begin{tabular}{@{}#1@{}}#2\end{tabular}}
\footnotesize
   \begin{tabular}{ r | c  |c|c|c|c}
     \hline
     \multirow{2}*{Method}  & \multirow{2}*{Data} & \multirow{2}*{Backbone} &  \multicolumn{2}{c|}{F-measure} & \multirow{2}*{FPS}\\
     \cline{4-5}
                            &                         &   & None & Full & \\
      \hline
     TextBoxes \cite{liao2017textboxes} & SynText800k, IC13, IC15, TT & ResNet-50-FPN & 36.3 & 48.9 & 1.4 \\
     \hline
     Mask TextSpotter'18 \cite{lyu2018mask} & SynText800k, IC13, IC15, TT & ResNet-50-FPN & 52.9 & 71.8 & 4.8 \\
     \hline
     Two-stage \cite{sun2018textnet} & SynText800k, IC13, IC15, TT & ResNet-50-SAM  & 45.0 & - & - \\
     \hline
     TextNet \cite{sun2018textnet} & SynText800k, IC13, IC15, TT & ResNet-50-SAM  & 54.0 & - & 2.7 \\
     \hline
     Li et al. \cite{li2019towards} & SynText840k, IC13, IC15, TT, MLT, AddF2k & ResNet-101-FPN & 57.80 & - & 1.4 \\
     \hline
     Mask TextSpotter'19
     \cite{liao2019mask} & SynText800k, IC13, IC15, TT, AddF2k & ResNet-50-FPN  & 65.3 & 77.4 & 2.0 \\
     \hline
     Qin et al. \cite{qin2019towards} & \tabincell{c}{SynText200k, IC15, COCO-Text, TT, MLT \\ Private: 30k (manual label), 1m (partial label)} & ResNet-50-MSF & 67.8 & - & 4.8 \\
     \hline
     CharNet \cite{xing2019convolutional} & SynText800k, IC15, MLT, TT & \tabincell{c}{ResNet-50-Hourglass57} & 66.2 & - & 1.2 \\
     \hline
     TextDragon \cite{feng2019textdragon} & SynText800k, IC15, TT & \tabincell{c}{VGG16} & 48.8 & 74.8 & - \\
     \hline
     \hline
      {\bf \BeCan-F}  & \multirow{3}*{SynText150k, COCO-Text, TT, MLT} & \multirow{3}*{ResNet-50-FPN} & 61.9 & 74.1 & {\bf 22.8} \\
      \cline{1-1}
      \cline{4-6}
      {\bf \BeCan}  &  &  & 64.2 & 75.7 & 17.9 \\
      \cline{1-1}
      \cline{4-6}
      {\bf \BeCan-MS}  &  &  & {\bf 69.5} & {\bf 78.4} & 6.9 \\
     \hline
   \end{tabular}
   \caption{Scene text spotting results on Total-Text.
   Here * represents the result is roughly inferred based on the original paper or the provided code. \BeCan-F is faster as the short size of input image is 600. MS: multi-scale testing. Datasets:
   AddF2k \cite{zhong2016deeptext};
   IC13   \cite{Karatzas2013ICDAR};
   IC15   \cite{karatzas2015icdar};
   TT     \cite{chng2019icdar2019};
   MLT    \cite{nayef2019icdar2019};
   COCO-Text \cite{veit2016coco}.
 }
   \label{tab:tt}
   \end{table*}

\subsection{Experimental results on Total-Text}

\paragraph{Dataset.} Total-text dataset \cite{kheng2017total} is one of the most important arbitrarily-shaped scene text benchmark proposed in 2017, which was collected form various scenes, including text-like scene complexity and low-contrast background. It contains 1,555 images, with 1,255 for training and 300 for testing. To resemble the real-world scenarios, most of the images of this dataset contain a large amount of regular text while guarantee that each image has at least one curved text. The text instance is annotated with polygon based on word-level. Its extended version \cite{ch2019total} improves its annotation of training set by annotating each text instance with a fixed ten points following text recognition sequence. The dataset contains English text only. To evaluate the end-to-end results, we follow the same metric as previous methods, which use F-measure to measure the word-accuracy.

\begin{figure}[!t]
   \centering
   \centerline{\includegraphics[width=8cm%
   ]
   {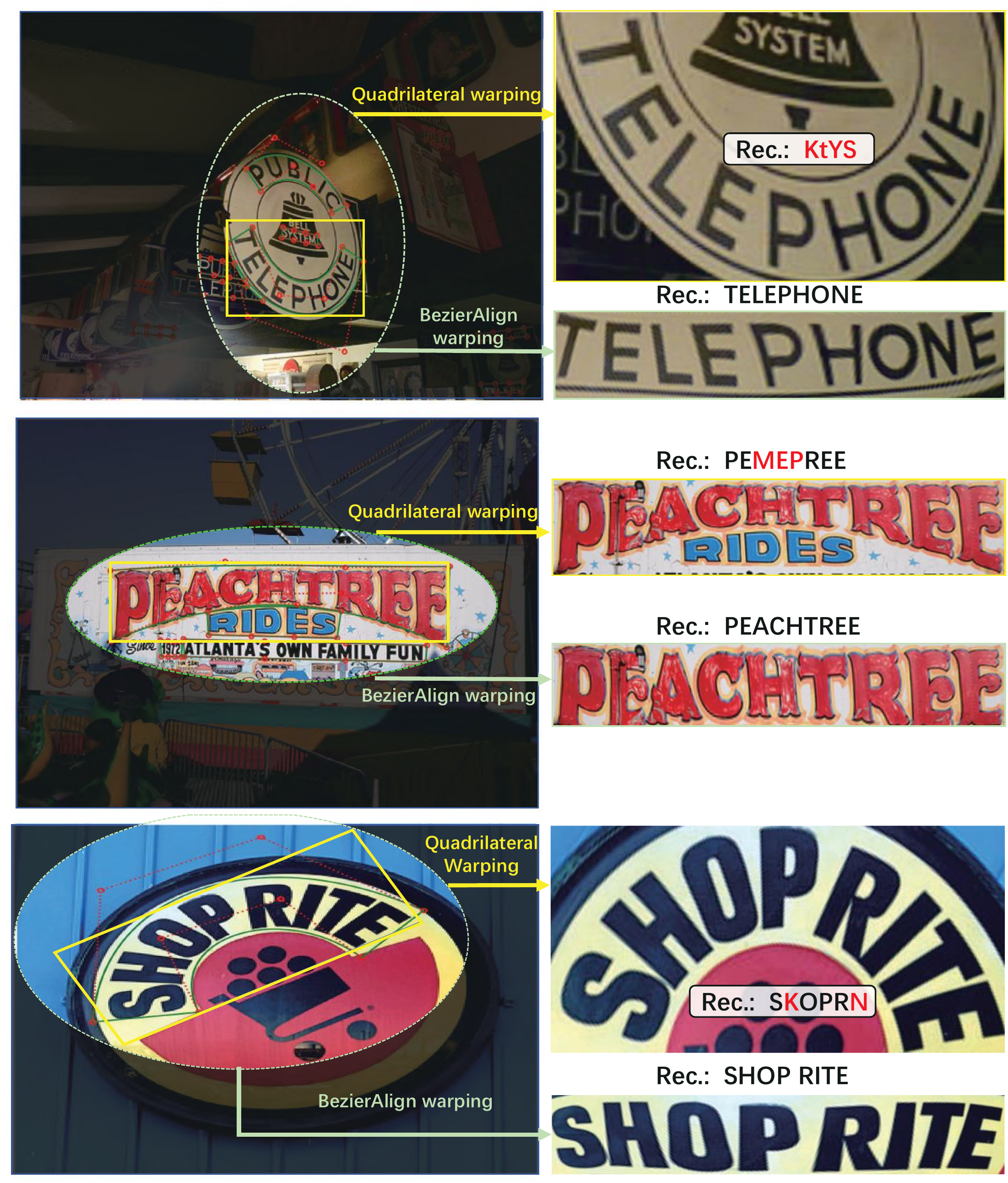}}
   \caption{Qualitative recognition results of the quadrilateral sampling method and BezierAlign. Left: original image.
   Top right:
   results by using quadrilateral sampling.
   Bottom right:
   results by using BezierAlign. }\label{fig:align_vis}
\end{figure}

\paragraph{Ablation studies: BezierAlign.} To evaluate the effectiveness of the proposed components, we conduct ablation studies on this dataset. We first conduct sensitivity analysis of how the number of the sampling points may affect the end-to-end results, which is shown in Table \ref{tab:exp_BezierAlign_sample_point}. From the results we can see that the number of sampling points can significantly affect the final performance and efficiency. We find (7,32) achieves the best trade-off between F-measure and FPS, which is used as the final setting in the following experiments. We further evaluate BezierAlign by comparing it with previous sampling method shown in Figure \ref{fig:Bezier_align}. The results shown in Table \ref{tab:exp_BezierAlign} demonstrate that the BezierAlign can
dramatically
improve the end-to-end results. Qualitative examples are shown in Figure \ref{fig:align_vis}.

\paragraph{Ablation studies: Bezier curve detection.} Another important component is Bezier curve detection, which enables arbitrarily-shaped scene text detection. Therefore, we also conduct experiments to evaluate the time consumption of Bezier curve detection. The result in Table \ref{tab:exp_Beziercurve_det} shows that the Bezier curve detection does not introduce extra computation compared with standard bounding box detection.
\begin{table}[!t]
   \centering
   \newcommand{\tabincell}[2]{\begin{tabular}{@{}#1@{}}#2\end{tabular}}
   \small
   \begin{tabular}{c| l |c}
     \hline
     Methods & Sampling method & F-measure (\%)  \\
     \hline
     \BeCan & with  Horizontal Sampling  & 38.4 \\
     \BeCan & with  Quadrilateral Sampling  & 44.7 \\
     \BeCan & with  BezierAlign  & 61.9 \\
     \hline
   \end{tabular}
   \caption{Ablation study for BezierAlign. Horizontal sampling follows \cite{li2017towards}, and quadrilateral sampling follows \cite{he2018end}. }
   \label{tab:exp_BezierAlign}
\end{table}

\begin{table}[!t]
   \centering
   \newcommand{\tabincell}[2]{\begin{tabular}{@{}#1@{}}#2\end{tabular}}
   \small
   \begin{tabular}{c|c|c|c}
     \hline
     Method & \tabincell{c}{Sampling points \\ ($n_h$, $n_w$)} & F-measure (\%) & FPS  \\
     \hline
     \multirow{6}*{\BeCan} & + (6, 32)  & 59.6 & {\bf 23.2} \\
     & + (7, 32)  & {\bf 61.9} & 22.8 \\
     & + (14, 64) & 58.1 & 19.9 \\
     & + (21, 96) & 54.8 & 18.0 \\
     & + (28, 128)  & 53.4 & 15.1 \\
     & + (30, 30)  & 59.9 & 21.4 \\
     \hline
   \end{tabular}
   \caption{Ablation study of the number of sampling points of BezierAlign.}
   \label{tab:exp_BezierAlign_sample_point}
\end{table}

\begin{table}[!t]
   \centering
   \newcommand{\tabincell}[2]{\begin{tabular}{@{}#1@{}}#2\end{tabular}}
   \small
   \begin{tabular}{c|c}
     \hline
     Methods & Inference time\\
     \hline
     without Bezier curve detection  & 22.8 fps \\
     with Bezier curve detection  & 22.5 fps \\
     \hline
   \end{tabular}
   \caption{Ablation study for time consumption of the Bezier curve detection. }
   \label{tab:exp_Beziercurve_det}
 \end{table}

\paragraph{Comparison with state-of-the-art.} We further compare our method to previous methods. From the Table \ref{tab:tt}, we can see that our single scale result (short size being 800) can achieve a competitive performance meanwhile
achieving
a real time inference speed, resulting in a better trade-off between speed and word-accuracy. With multi-scale inference, \BeCan achieves state-of-the-art performance, significantly outperforming all previous methods especially in the running time. It is worth mentioning that our faster version can be more than 11 times faster than previous best method \cite{liao2019mask} with
\textit{on par} accuracy.

\begin{figure*}[!t]
  \centering
  \centerline{\includegraphics[width=17.6cm, height = 8.0cm]{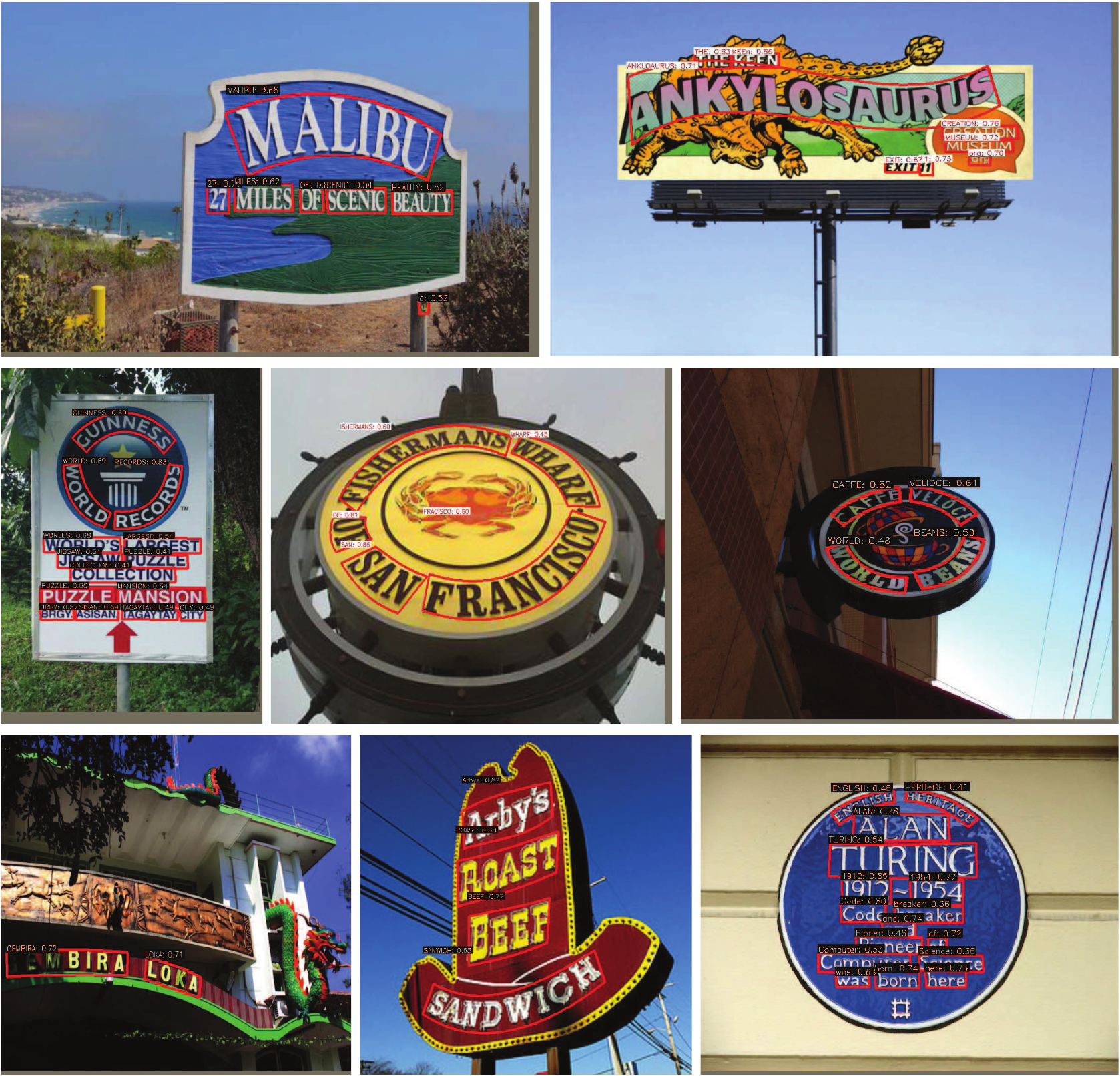}}
  \caption{Qualitative results of \BeCan on the Total-text. The detection results are shown with red bounding boxes.
   The float number is the predicted confidence. Zoom in for better visualization. }\label{fig:final}
\end{figure*}

\paragraph{Qualitative Results.} Some qualitative results of \BeCan are shown in Figure \ref{fig:final}. The results show that our method can accurately detect and recognize most of the arbitrarily-shaped text. In addition, our method can also well handle straight text, with nearly quadrilateral compact bounding box and correct recognize results. Some errors are also visualized in the figure, which are mainly caused by mistakenly recognizing one of the characters.

\subsection{Experimental Results on CTW1500}
\paragraph{Dataset.} CTW1500 \cite{liu2019curved} is another important arbitrarily-shaped scene text benchmark proposed in 2017. Compared to Total-Text, this dataset contains both English and Chinese text.
In addition, the annotation is based on text-line level, and it also includes some document-like text, i.e., numerous small text may stack together. CTW1500 contains 1k training images, and 500 testing images.

\begin{figure}[!t]
  \centering
  \centerline{\includegraphics[width=8.6cm]{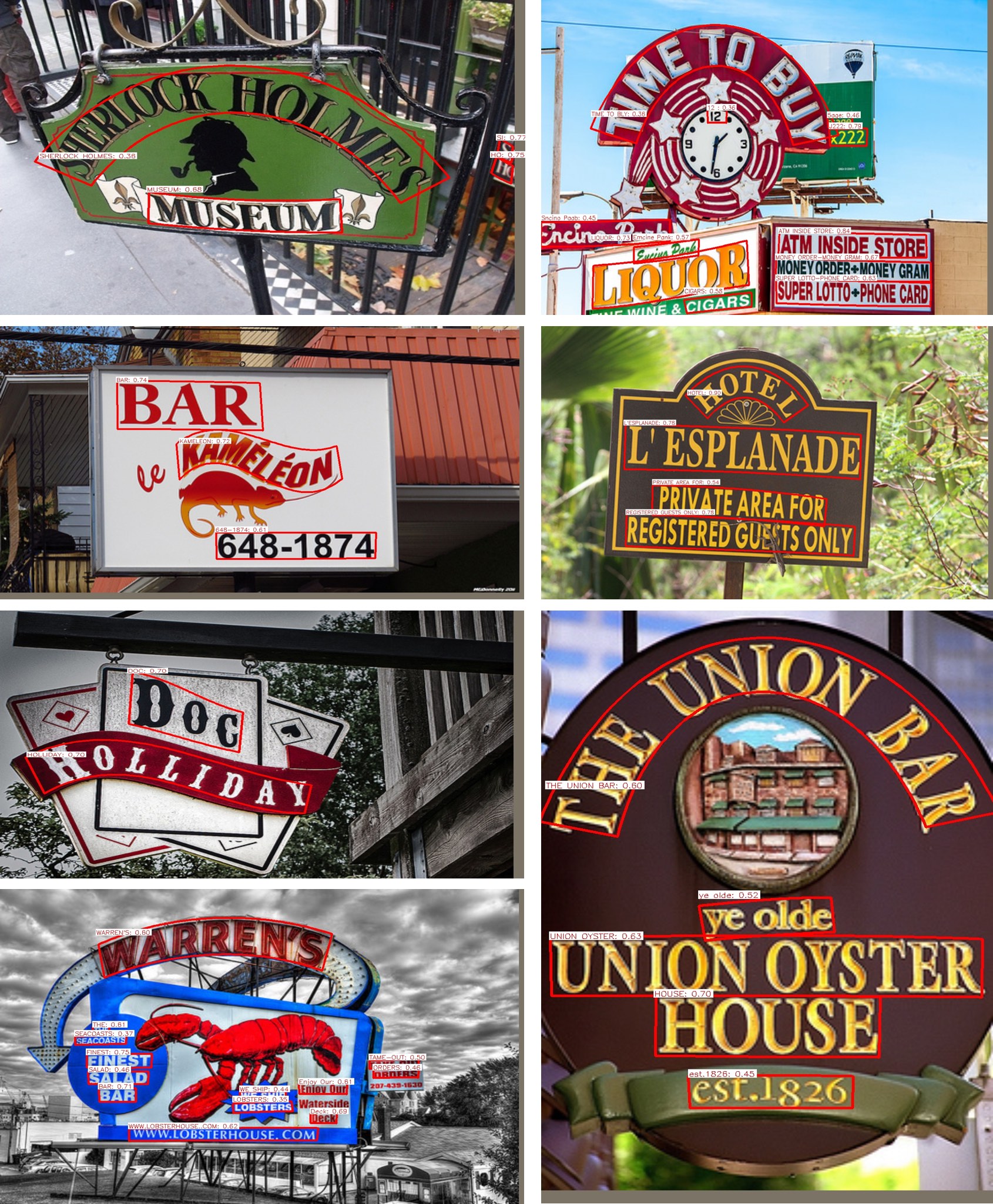}}
  \caption{Qualitative end-to-end spotting results of CTW1500. Zoom in for better visualization.}\label{fig:ctw1500}
\end{figure}

\paragraph{Experiments.} Because the occupation of Chinese text in this dataset is very small, we directly regard all the Chinese text as ``unseen'' class during training, i.e., the 96-$th$ class. Note that the last class, i.e., the 97$^{th}$ class is ``EOF'' in our implementation. We follow the same evaluation metric as \cite{feng2019textdragon}. The experimental results are shown in Table \ref{tab:exp_ctw1500}, which demonstrate that in terms of end-to-end scene text spotting, the \BeCan can significantly surpass previous state-of-the-art methods. Examples results of this dataset are showed in Figure \ref{fig:ctw1500}.
From the figure, we can see that some long text-line instances contain many words, which make a full-match word-accuracy extremely difficult.
In other words mistakenly recognizing one character will result in zero score for the whole text.

\begin{table}[!t]
  \centering
  \newcommand{\tabincell}[2]{\begin{tabular}{@{}#1@{}}#2\end{tabular}}
  \footnotesize
  \begin{tabular}{r |c|c|c}
    \hline
    \multirow{2}*{Methods} & \multirow{2}*{Data} & \multicolumn{2}{c}{ F-measure}\\
    \cline{3-4}
    & & None & Strong Full \\
    \hline
    FOTS \cite{liu2018fots} & SynText800k, CTW1500 & 21.1 & 39.7 \\
    \hline
    Two-Stage* \cite{feng2019textdragon} & SynText800k, CTW1500 & 37.2 & 69.9\\
    \hline
    RoIRotate* \cite{feng2019textdragon} & SynText800k, CTW1500 & 38.6 & 70.9\\
    \hline
    LSTM* \cite{feng2019textdragon} & SynText800k, CTW1500 & 39.2 & 71.5\\
    \hline
    TextDragon \cite{feng2019textdragon} & SynText800k, CTW1500 & 39.7 & 72.4\\
    \hline
    \hline
    \BeCan & SynText150k, CTW1500 & {\bf 45.2}  & {\bf 74.1} \\
    \hline
  \end{tabular}
  \caption{End-to-end scene text spotting results on CTW1500. * represents the results are from \cite{feng2019textdragon}. ``None'' represents lexicon-free. ``Strong Full'' represents that we  use all the words appeared in the test set. }
  \label{tab:exp_ctw1500}
\end{table}

\section{Conclusion}
We have proposed \BeCan---a real-time end-to-end method that uses Bezier curves for arbitrarily-shaped scene text spotting.
By reformulating arbitrarily-shaped scene text using parameterized Bezier curves, \BeCan can detect arbitrarily-shaped scene text with Bezier curves which  introduces negligible computation cost compared with standard bounding box detection. With such regular Bezier curve bounding boxes, we can naturally connect a light-weight recognition branch via a new BezierAlign layer.

In addition, by using our Bezier curve synthesized dataset and publicly available data, experiments on two arbitrarily-shaped scene text benchmarks (Total-Text and CTW1500) demonstrate that our \BeCan can achieve state-of-the-art performance, which is also significantly faster than previous methods.

\section*{Acknowledgements}
The authors would like to thank
Huawei Technologies for the donation of
GPU cloud computing resources.

{\small
\bibliographystyle{ieee_fullname}
\bibliography{Bezier}
}

\end{document}